\documentclass[10pt,twocolumn,letterpaper]{article}
\usepackage{times}
\usepackage{epsfig}
\usepackage{graphicx}
\usepackage{amsmath}
\usepackage{amssymb}
\usepackage{cite}
\usepackage[colorlinks,linkcolor=blue,anchorcolor=blue,citecolor=red,]{hyperref}
\usepackage{booktabs}
\usepackage{float}
\usepackage{multicol}
\usepackage[misc]{ifsym}

\begin{document}
	\title{Using Adaptive Gradient for Texture Learning  in Single-View 3D Reconstruction}
	\author{
		\begin{tabular}{c} Luoyang Lin \\ Shenzhen Institute of Artificial \\Intelligence and Robotics for Society\\ 	{\tt\small linluoyang@cuhk.edu.cn} \end{tabular} \and
		\begin{tabular}{c} Dihong Tian \\ Shenzhen Institute of Artificial \\Intelligence and Robotics for Society \\ 	{\tt\small tiandihong@cuhk.edu.cn} \end{tabular}

	}
	\date{}
	\maketitle
\begin{abstract}
	Recently, learning-based approaches for 3D model reconstruction have attracted attention owing to its modern applications such as Extended Reality(XR), robotics and self-driving cars. Several approaches presented good performance on reconstructing 3D shapes by learning solely from images, i.e., without using 3D models in training. Challenges, however, remain in texture generation due to the gap between 2D and 3D modals. In previous work, the grid sampling mechanism from Spatial Transformer Networks was adopted to sample color from an input image to formulate texture. Despite its success, the existing framework has limitations on searching scope in sampling, resulting in flaws in generated texture and consequentially on rendered 3D models. In this paper, to solve that issue, we present a novel sampling algorithm by optimizing the gradient of predicted coordinates based on the variance on the sampling image. Taking into account the semantics of the image, we adopt Fr$\acute{e}$chet Inception Distance (FID) to form a loss function in learning, which helps bridging the gap between rendered images and input images. As a result, we greatly improve generated texture. Furthermore, to optimize 3D shape reconstruction and to accelerate convergence at training, we adopt part segmentation and template learning in our model. Without any 3D supervision in learning, and with only a collection of single-view 2D images, the shape and texture learned by our model outperform those from previous work. We demonstrate the performance with experimental results on a publically available dataset. 
	
\end{abstract}
\section{Introduction}

\begin{figure}
	\includegraphics[width=1.0\linewidth]{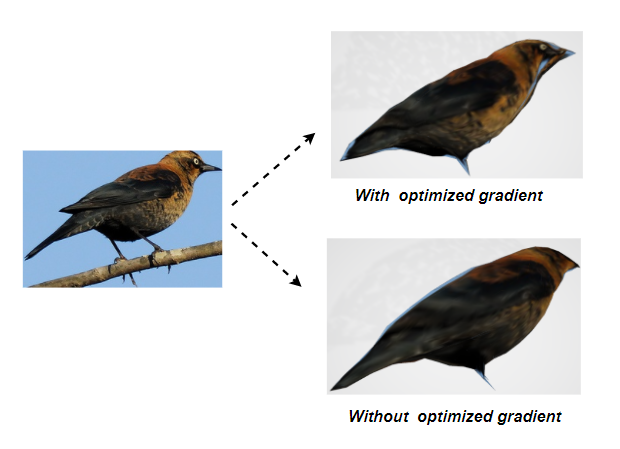}
	
	\caption{(a) 2D input image, (b) Rendered image by our model with optimized gradient. (c) is the rendered image learned by UMR\cite{li2020self} without optimized gradient. As showed in Figure \ref{intro_png}, the model with optimized gradient can sample more reasonable coordinates from input image for texture prediction, e.g. the texture for the eye of the bird. }
\label{intro_png}
	
\end{figure}

\indent Three Dimensional (3D) reconstruction from single-view images is an ill-posed problem and a big challenge in computer vision. It aims to recover the shape and texture of a 3D model from a single image. Previous methods are divided into several families based on the use of supervisory information in learning. Some methods\cite{achlioptas2018learning,wu2018learning,wang2018pixel2mesh} tackled the problem by utilizing ground-truth 3D shapes, i.e., using explicit 3D information to supervise learning. Without explicit 3D supervision, Kanazawa\cite{kanazawa2018learning} reconstructs 3D shape and texture by leveraging semantic keypoints on rendered 2D images. More recent approaches\cite{kato2019learning,xie2019pix2vox,wang20193dn} achieve 3D reconstruction only based on 2D images, with multi-view images for each object instance. The common drawback of the aforementioned approaches is the cost of aquiring training data. Both ground-truth 3D models and multi-view images per object are difficult to obtain. To alleviate the issue, Liu\cite{liu2019soft} and Kanazawa\cite{kanazawa2018learning} proposed methods to learn 3D reconstruction models only with 2D supervision and a single image for each object instance. Their methods are sucessful in reconstructing 3D shapes, but at a loss of fidelity in texture. Inspired by their work, in this paper, we propose a novel framework to optimize both shape and texture generation in 3D reconstruction, using single image supervision.
\\
\indent For sole 2D supervision with a single-view image, previous work relies on input images and silhouettes to predict the texture for the 3D model. By matching the rendered image and predicted boundary with the input 2D image and its silhouette, previous work focus on learning 3D shape and texture from 2D information, but it often falls into model collapse due to a phenomenon called camera-shape ambiguity. Although the generated 3D shape is different from the real 3D object, the incorrect camera pose can lead to little difference between rendered and input 2D images, and gradient will decay close to 0, resulting in collapse of the model. Li\cite{li2020self} has put forward semantic part invariance to eliminate that ambiguity. It leverages semantic image segmentation to adjust the predicted camera pose and hence improves shape learning to some extent. Goel\cite{goel2020shape} adopted a decoder to predict the texture directly rather than sampling from the input image to optimize camera poses. Their work gives full consideration on shape and camera-pose optimization but not as much on texture generation. 
\\
\indent In this paper,  inspired by CMR\cite{li2020self}, we propose a novel framework to learn the shape and texture of a 3D model from a single-view image. For 3D shape and camera pose learning, we utilize semantic image segmentation and template learning to optimize 3D coordinates of vertices and the camera pose for an input image.
 For shape generation, we update the template to possess richer object detail.
 Through the image segmentation, we leverage UMR\cite{li2020self} to obtain vertices for every part of each object in the training to maintain the shape learning.
 For texture generation, the texture flow is learned to sample pixels from the sampling image to 3D mesh, following the differentiable sampling framework, $Grid\_Sample$, as described in Jaderberg\cite{jaderberg2015spatial}. To improve sampling, we propose a new algorithm to optimize gradient by modifying the distribution of the sampling image. As shown in Figure \ref{intro_png}, with the optimized gradient, the texture flow can search the more reasonable region in the texture coordinate space, and the resulting object possesses more detail such as eyes and body. To further bridge the gap between 2D and 3D modal, we include Fr$\acute{e}$chet Inception Distance (FID)\cite{heusel2017gans} as a loss function to enforce the consistency between distribution of input images and rendered images. To the best of our knowledge, our work is the first to introduce FID into the field of single-view 3D reconstruction.

Our main contribution in this paper is two fold:

(1) For texture learning, we propose a novel sampling algorithm to optimize the gradient of predicted coordinates when sampling for texture.

 
(2) For both shape and texture learning, we introduce FID as a loss function to bridge the gap between rendered images and input images from the perspective of distribution.

\section{Related work}

\indent High fidelity 3D model reconstruction has received a lot of attention from Virtual Reality (VR), Augmented Reality (AR), robotics and so on. Given an input such as a single image, multiple images or a point could, 3D reconstruction aims to recover realistic shape and texture that resemble the input. Various learning-based algorithms have been proposed in previous work, using different degrees of supervision such as ground-truth 3D shape, multi-view images, and single-view images.

\textbf{Reconstruction based on 3D supervision or multi-view images:} Previous methods \cite{achlioptas2018learning,wu2018learning,wang2018pixel2mesh,wen2019pixel2mesh++,fan2017point,gkioxari2019mesh} reconstruct 3D models based on 3D supervisory information, i.e. the coordinates of each vertex in 3D shape, enabled by synthetic datasets such as SunCG\cite{song2017semantic} and ShapeNet\cite{wu20153d}. Chamfer distance and Earth Mover distance are always adopted to measure the distance between a predicted point set and the ground-truth set for shape learning. One drawback of those distance measures is that they can not guarantee the predicted points are on the surface of the ground-truth set. In response to it, Jiang\cite{jiang2018gal} proposed Geometric Adversarial Loss, including Multi-view Geometric Loss and Point-based Conditional Adversarial Loss, to constrain the position of generated points based on images from various perspectives and global features extracted by Qi\cite{qi2017pointnet}. Methods \cite{qi2017pointnet,qi2017pointnet++} were adopted to extract global features to maintain 3D shape learning.

Obviously, 3D supervision requires abundant pairs of a single image and its corresponding 3D shape, which requires significant manual effort to annotate. To avoid 3D supervision, a few researchers \cite{kato2019learning,xie2019pix2vox,wang20193dn,yu2020fast,yang2018learning,liu2019learning,gadelha20173d} proposed algorithms for 3D reconstruction based on multi-view images. These methods predict 3D shape through the correspondence between 2D and 3D modal realized, for example, by the differentiable renderer proposed by Kato\cite{kato2018neural}.

\textbf{Reconstruction based on retrieval and deformation:} To generate 3D shape such that it resembles the target shape as close as possible, previous work\cite{uy2020deformation,uy2021joint,wang20193dn} combined retrieval with deformation for 3D shape reconstruction. Given a query, e.g. an input image or a point cloud, the model will retrieve an existing shape in a collected 3D dataset and deform it to satisfy the input query. Apparently, it is a high price to establish such a 3D model dataset. 

\textbf{Reconstruction based on single-view images:} Realistically, multi-view images from abundant object instances are difficult to acquire. For that reason, recent work like Salvi\cite{salvi2020attention} introduced learning-based architectures for 3D shape reconstruction using solely single input images. Thanks to the differentiable renderer\cite{chen2019learning,kato2018neural,liu2019soft}, several frameworks \cite{goel2020shape,li2020self,chen2019learning,kato2018neural,liu2019soft,kanazawa2018learning} were presented to bridge the gap between an input image and its resulting texture by utilizing differentiable rendering and image reconstruction. In those frameworks, the reconstruction task was generally decomposed into three modules, shape, texture and camera pose learning, respectively. There are primarily two approaches for texture generation, sampling from input images or direct color prediction by neural network. With respect to camera poses, Kanazawa\cite{kanazawa2018learning} utilized annotated 2D keypoints to obtain more precise camera poses.

\section{Proposed Method}
We propose a new framework to predict a 3D model from a single-view image. Given a 2D input image $I_r$, our framework first leverages Resnet\cite{he2016deep} as the main framework to extract features from the image, denoted as $I_f$. Then, the framework predicts the 3D mesh shape, texture, and camera pose by three decoders, $D_{shape}, D_{flow}, D_{cam}$, respectively, similar to those in UMR\cite{li2020self}. Different from UMR\cite{li2020self}, however, to optimize gradient in sampling, we also introduce a variance learning decoder, $D_{var}$, to apply variation to the sampling image. 

For shape reconstruction, we adopt the Intersection-Over-Union (IOU) loss\cite{chen2019learning}, a distance function between a predicted silhouette $S_{pred}$ and the input silhouette $S_{real}$. We additionally leverage RGB information to conduct 3D shape prediction. One intuitive approach is to keep consistency between the predicted view from a predicted camera pose and the original image. Besides the view from the input image, we can optimize the 3D shape from the various views. Obviously, the images from other views are absence and one can not adopt the image reconstruction directly to restrain the shape prediction. Fortunately, GAN (Generative Adversarial Network)\cite{goodfellow2014generative} provides a weak supervision to enhance the unseen part learning. 

\begin{figure}[h]
	\includegraphics[width=1.0\linewidth]{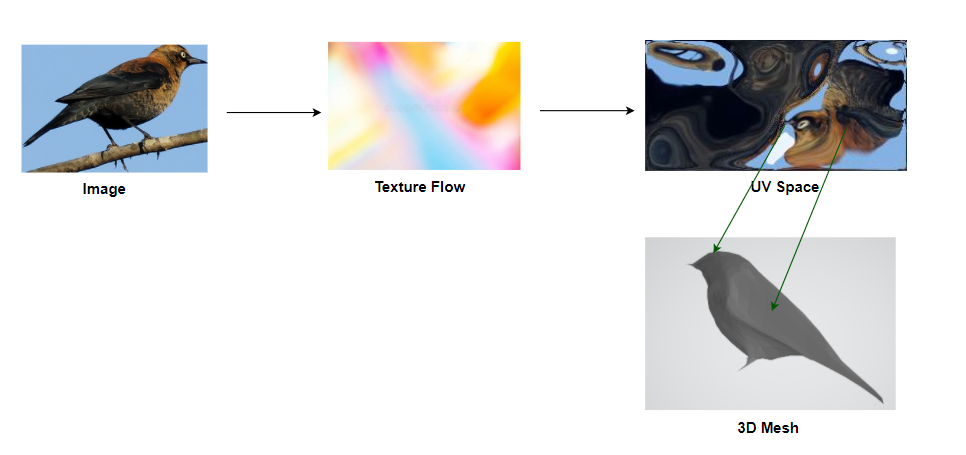}
	
	\caption{As showed in Figure \ref{img_flow_uv_png}, the decoder $D_{flow}$ generate the $I_{flow}$ to project the pixel from 2D input image into UV space as a mapping
function and pre-defined mapping function $\phi$ maps the color in UV space into each vertex of 3D mesh for texture generation. }
\label{img_flow_uv_png}
	
\end{figure}

For texture generation, we consider it as a mapping from the input image to the texture coordinate space, a.k.a. the UV space. More specifically, the decoder $D_{flow}$ is learned for predicting the coordinates $I_{flow}$ for sampling and the $I_{flow}$ projects the color from input image into the UV space, $I_{uv}$. Then the pre-defined function $\phi$ establishes the mapping between the UV space and 3D mesh for generating the texture. The entire process can be illustrated by Figure \ref{img_flow_uv_png}. Due to the limitation of gradient in sampling, we propose a novel algorithm for predicting coordinates, that is, we adopt a decoder $D_{var}$ to apply the variance into the sampling image to optimize the gradient of coordinates.
To learn the parameters for $D_{flow}$ and $D_{var}$, image reconstruction between the input image and rendered image is adopted to establish the mapping between the sampling image and $I_{uv}$.
We additionally leverage the pseudo-label provided by SoftRas to optimize coordinates learning.

For the generation of unseen part, FID is adopted as a loss function to encourage the shape and texture learning at semantic level.

To accelerate the convergence of training, we divide model training into two stages. In the first stage, we train a template for deformation and learn vertex segmentation from a sphere as UMR. Based on the learned template,  we train 3D coordinates of mesh by deformation and enforce constraint on each part with vertex segmentation. To optimize the shape and texture in a greater
degree, we propose an adaptive algorithm and adopt FID as a loss function in our model. 

\subsection{Basic Architecture}

\indent In this section, we introduce our basic framework for 3D model generation including shape, camera pose and texture prediction. This is done by following the decoders $D_{shape},D_{flow},D_{cam}$ and sharing the feature extracted by the encoder $F_e$. Different from previous work, we have proposed a novel variance decoder $D_{var}$ for gradient optimization. For 3D shape reconstruction, since ground-truth 3D shape or other keypoints information are not included in our training, the prediction for 3D shape only relies on 2D images and corresponding silhouettes from our collection.

Intersection-Over-Union(IOU) is adopted as a loss function to train the decoders $D_{shape}$ and $D_{cam}$ with an differentiable renderer. For silhouette reconstruction:
\begin{align}
	L_{iou}=E_{\theta}
[ 1- \frac{ \|S_{real} \bigodot S_{pred}\|_{1} }{\| S_{real}+S_{pred}- S_{real} \bigodot S_{pred} \|_{1}}  ]
\label{eq:iou}
\end{align} 
where $ \bigodot, S_{real} $ denote element-wise product and silhouettes extracted from input images. The predicted silhouettes,  $S_{pred}$, in Equation\eqref{eq:iou}, 

\begin{align}
	S_{pred}=F_r(D_{shape}(F_e(I_{r}))+V_{temp}, D_{cam}(F_e(I_r)))
	~\label{eq:dshape}
\end{align}
$F_e, F_r, I_r$ are the feature extractor for image, differentiable rendering and input images, respectively. In Equation~\eqref{eq:dshape}, 3D shape prediction based on template deformation is included to learn shape stably. $D_{shape}$ aims to learn an offset from pretrained template $V_{temp}$ to a 3D shape which satisfies the input image.


Besides silhouette reconstruction, we add three extra regularization terms to optimize the 3D shape. First, we adopt a regularization term to prevent excessive deformation\cite{kar2015category}. 
\begin{align}
	L_{\triangle V} =\| \delta V \|=\| D_{shape}(F_e(I_r)) \| 
\end{align}
In addition, a Laplacian term is included to enforce the smoothness of 3D Mesh~\cite{liu2019softeq}.
\begin{align}
	L_{lap} &=\sum_{p} \| \delta_{p} \| \\
		\delta_{p}&=v_p- \frac{1}{|N(p)|} \sum_{q \in N(p)}v_{q},
\end{align}
where $v_p$ is a vertex of the mesh and $N(p)$ is the neighborhood vertices of p. $L_{lap}$ aims to prevent the vertices from moving too freely. Last, we adopt  $L_{flat}$ to encourage the smoothness of surface and prevent self-intersections~\cite{liu2019softeq},
\begin{align}
	L_{flat}=\sum_{\nu \in \{\nu_i\} } \| cos(\nu)+1 \|,
\end{align}
 where $\nu_i$ is the angle between two faces which share a common edge. When they stay on the same plane, $L_{flat}$ reaches its minimum value.

For texture generation, we adopt a mode $Image-UV Space-Tex$ to formula textures.
First the decoder $D_{flow}$ and $Grid\_Sample$  are included to map pixels from the 2D input image into UV space $I_{uv}$. 
\begin{align}
	I_{uv}&=GS(D_{flow}(F_e(I_r)),I_r)\\
	&=GS(I_{flow},I_r)	
\end{align}
The texture flow $I_{flow}$ is the output of $D_{flow}$ and  establishes an mapping from the input image to UV space. Then,
 a pre-defined mapping function $\phi$ maps pixels from the UV space into points of 3D mesh to formulate textures $Tex$.
\begin{align}
	Tex= \phi(I_{uv})
\end{align}

Similar to UMR\cite{li2020self}, image reconstruction is adopted to train the texture flow $I_{flow}$. Formally,
\begin{align}
	L_{rec}=E_{\theta}[\| I_r-F_r(V_{pred},\phi(GS(I_{flow},I_r)),cam) \|]	
\end{align}
$V_{pred}$ is the learned shape by $D_{shape}$ and $V_{temp}$. There is, however, still a flaw with $GS$ in image reconstruction due to the limitation for its sampling mechanism. To be more precise, predicted coordinates fall into local region and can not be updated accurately according to
image reconstruction due to its invalid gradient.

 To alleviate the issue, we propose a novel algorithm based on variance applied to sampling images. In our framework, an variance decoder $D_{var}$ is proposed to learn an variance $I_{var}$ for sampling images. 
$I_{var}$ aims to optimize the gradient for predicted coordinates to search reasonable region(More detail can be found in Section \textbf{3.2}). Following $I_{var}$, $I_{uv}$ is updated as follows,
\begin{align}
	I_{uv}&=GS(I_{flow},I_r*I_{var})
\end{align}
In Section \textbf{3.2}, we will introduce the gradient optimization with $I_{flow}$ and  $I_{var}$ in detail.

Besides image reconstruction, we leverage a pseudo-label provided by Softras to optimize the texture flow, which provided general location for predicted coordinates. More specifically, considering all points on a face $F_0$ of 3D mesh, there is a weak constraint between the texture flow and the pseudo-label according to Liu\cite{liu2019soft}. 
 In formula,
\begin{align}
	L_{align}=\| \frac{1}{n_c}  \sum_{x \in F_0 } I_{flow}^{-1} (\phi^{-1}(x))  -   \frac{1}{n_c} \sum_{i,j} w_{ij}c_{ij}     \|
	\label{fn_align}	
\end{align}
$w_{ij}, c_{ij}$ denote the weight of $F_0$ to the pixel in the image provided by SoftRas and the normalized coordinate of the image
at position $i,j$. $n_c$ is the number of points on $F_0$.

At a semantic level, a perceptual distance\cite{zhang2018unreasonable} is adopted to bridge the gap between predicted images and input images for texture learning, leveraging the latent feature extracted by VGG\cite{simonyan2014very}.

To train the camera pose in a greater degree, vertex segmentation and pretrained template are adopted to optimize $D_{cam}$\cite{li2020self}. For vertices in each segmented part of pretrained template, we project 3D coordinates
into 2D space with predicted camera pose and calculate a distance between them and part label $P^{i}_{uv}$. 
$P^{i}_{uv}$ is a coordinate estimate  calculated by Hung\cite{hung2019scops}. For the distance,
\begin{align}
	L_{part}=chamfer( \psi(V_i) ,P^{i}_{uv})
\end{align}
$V_i$ is a set of i-th part vertices of template and $\psi$ is a mapping function to project vertices from 3D space into 2D space according to predicted camera pose. Chamfer loss is adopted to calculate the distance between 2D projected points and the label $P_{uv}^i$. 

With the part segmentation learned by UMR, probability prediction is adopted to optimize $D_{cam}$\cite{li2020self}. 
\begin{align}
	L_{prob}=\|P_r-F_r(V_{seg},cam)\|
\end{align} 
$P_{r}$ denotes a part segmentation probability map and the $V_{seg}$ is the segmentation for vertices on 3D mesh, pretrained by UMR.

\subsection{Adaptive Gradient Based on Variance}

\begin{figure}[h]
 	\centering
 	\includegraphics[width=0.85\linewidth]{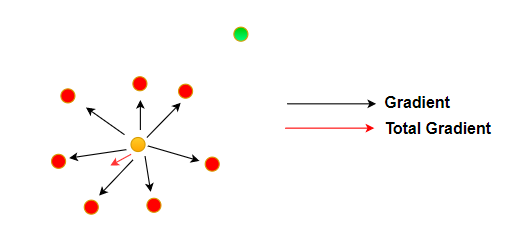}
 	
 	\caption{As showed in Figure \ref{grad0_png}, the yellow point is current predicted coordinate and it can only accept the gradient
 		from  the red points surround it due to limitation of $Grid\_Sample$. Total gradient from red points is small and discourages
 		the yellow point move forward to the target position, the green point.} 
 	\label{grad0_png}
 \end{figure}

\indent In our model, $D_{flow}$ and $Grid\_Sample$ are adopted to establish a mapping from input images and UV space.
With $Grid\_Sample$, one can sample the pixels from input images and the whole process is differentiable.
$Grid\_Sample$, however, only captures a surrounding tiny area and causes an limitation to parameters optimization. 
Due to the limitation, coordinates updating falls into a collapse when training with $L_{rec}$ and $L_{align}$. 
A plain example is showed in Figure \ref{grad0_png}.

\begin{figure}[h]
	\centering
	\includegraphics[width=0.85\linewidth]{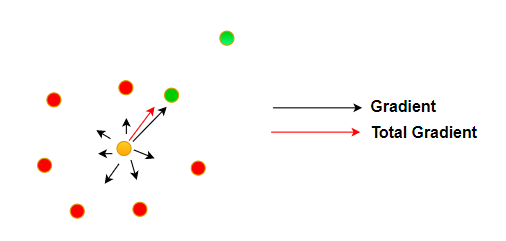}
	
	\caption{As expressed in Figure \ref{grad1_png}, after adjusting distribution of surrounding points of current predicted coordinate(the yellow point), the 
		gradient from red points comes to weak and the gradient from the green point(the nearest point to the yellow point)
		takes a leading role. Total gradient enforce the current predicted point moves forward the target position, the green point(the farthest point away from the yellow point).
	} 
\label{grad1_png}
\end{figure}

 As illustrated in Figure \ref{grad0_png}, the yellow point can not catch reasonable region because there is no valid gradient from its neighborhood, a few surrounding points. As a result, coordinates update and texture learning will fall into a collapse.
 To alleviate the issue,
 an intuitive solution is to modify color distribution of surrounding points. The optimization can be expressed frankly in Figure \ref{grad1_png}.
 
\begin{figure*}[ht]
	\centering
	\includegraphics[width=1.0\linewidth]{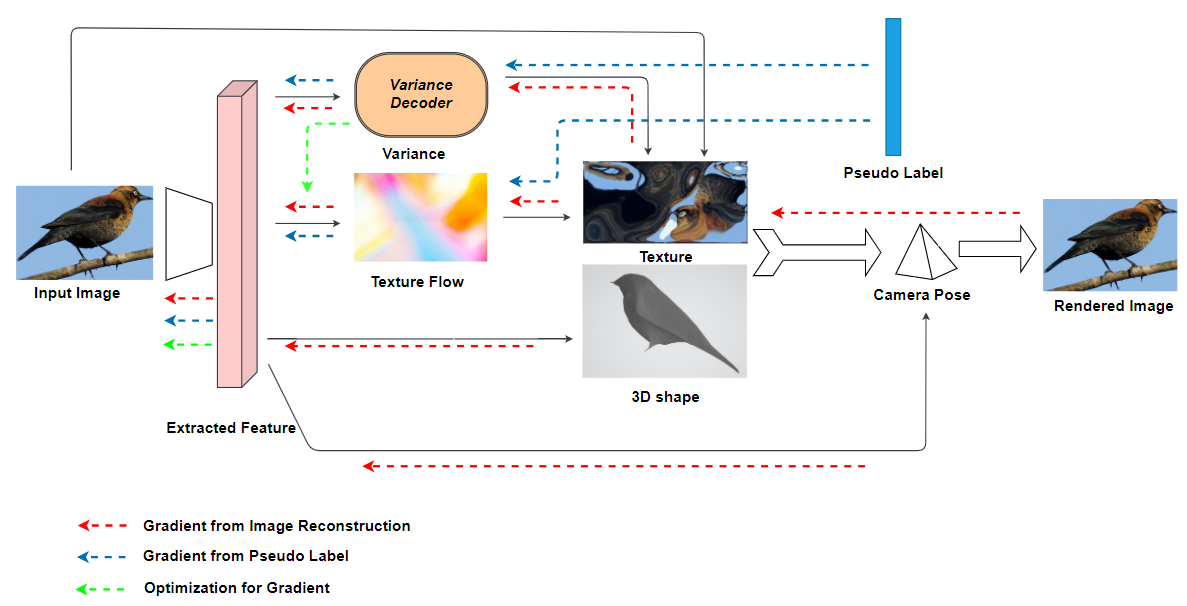}
	
	\caption{As expressed in Figure \ref{arch_png}, our basic architecture is composed of an encoder $F_e$ and several decoders
		including $D_{shape}, D_{flow}$, $D_{cam}$ and $D_{var}$. $F_e$ aims to extract an sharing feature for decoders.
		$D_{shape}, D_{flow}$ and $D_{cam}$ are trained to learn 3D shape, the texture flow, and the camera pose for 3D model.
		To adjust the gradient for the texture flow, $D_{var}$ is adopted to apply a variance for sampling images, which is showed
		by the green dashed line. The gradient of predicted coordinates mainly comes from the pseudo label and Image reconstruction, correspond to the blue and red dashed line,  respectively.    
	} 
\label{arch_png}
\end{figure*}

Taking the perspective of neural network learning, parameters for sampling coordinates can not search for a reasonable region because 
there is no invalid gradient for updating.
Inspired by Figure \ref{grad1_png}, we propose a novel sampling algorithm to optimize the gradient for texture flow $D_{flow}$.

\begin{figure}[ht]
	\centering
	\includegraphics[width=1.0\linewidth]{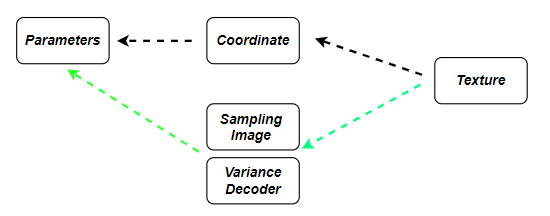}
	
	\caption{As showed in Figure \ref{sim_flow_png}, besides original gradient in the training(the black dotted arrow), we introduce a new branch(the green dotted arrow) to optimize parameters based on a variance applied to sampling images.
	} 
\label{sim_flow_png}
\end{figure}

For sampling from input images to UV space, $Grid\_Sample$ is still included in our algorithm to formula textures.
We, however, raise a new branch for gradient optimizing in the process of sampling.
A simplified version is showed in Figure \ref{sim_flow_png}.
In previous work, the gradient to optimize coordinates only from textures, no considering the input image. In formula,
\begin{align}
	I^{old}_{grad}&=\frac{\partial}{\partial \theta_{c}} Grid\_Sample(I,\theta_{c})=h(I,\theta_{c})
\end{align}
$\theta_{c}$ is learned parameters for coordinates. Obviously, the gradient from textures is associated with sampling images $I$. In this paper, we introduce a new auxiliary gradient branch texture-sampling image-$\theta_{c}$ for back-propagation
following a variance decoder $D_{var}$. In particular, we adopt $D_{var}$ to apply a variance to sampling images from $I$ to $I'$. The gradient is adjusted as follows.

\begin{align}
	I^{new}_{grad}&=\frac{\partial}{\partial \theta_{c}} Grid\_Sample(I',\theta_{c})=h(I',\theta_{c})
\end{align}

In the process of sampling, $Grid\_Sample$ mainly depends on a multiplication operator. So the learned decoder $D_{var}$ applies the variance $I_{var}$ to the input image based on multiplying. The new gradient is calculated with $I_{var}$ as follows.
\begin{align}
	I_{grad}^{new}=\frac{\partial}{\partial \theta_{c}} Grid\_Sample(I_r*I_{var},\theta_{c})
	\label{grad_new_fn}
\end{align}
Based on $I_{var}$, the optimized gradient $I_{grad}^{new}$ prevents current predicted coordinates from falling into a collapse.
The whole optimizing process can be analyzed in formula as follows.

For original method of sampling, the output of sampling $C_{xy}$ at current position $x,y$ is calculated as follows.
\begin{align}
	C_{xy}=\sum_i \sum_j U_{ij}max(0,1-| x- i|)max(0,1-|y-j|)
	\label{gs_fn}
\end{align}
$U_{ij}$ is the pixel of sampling image at position $i,j$. To simplify the problem in the training, Suppose $C'_{xy}$ is the supervised label for $C_{xy}$ in image reconstruction. For image reconstruction at position $x,y$,
\begin{align}
	C^{rec}_{xy}=(C_{xy}-C'_{xy})^2
	\label{rec_xy}
\end{align}
The partial derivative of $C^{rec}_{xy}$ with respect to x is  calculated as follows
\begin{align}
	\frac{\partial C^{rec}_{xy}} {\partial x}=2(C_{xy}-C'_{xy}) \frac{\partial C_{xy}} {\partial x}
	\label{gradx}
\end{align}
Similarly, the partial derivative of $C^{rec}_{xy}$ with respect to $y$ is calculated as follows
\begin{align}
	\frac{\partial C^{rec}_{xy}} {\partial y}=2(C_{xy}-C'_{xy}) \frac{\partial C_{xy}} {\partial y}
	\label{grady}
\end{align}
According to \eqref{gs_fn}, obviously, valid
gradient for current coordinates $x,y$ is restricted to the interval $[x-1,x+1] \times [y-1,y+1]$, a tiny region discussed above.
Define $\lambda_{ij}$ as 
\begin{align}
	\lambda_{ij}=max(0,1-| x- i|)max(0,1-|y-j|)
	\label{lambda_ij}
\end{align}
The equation\eqref{rec_xy} can be rewritten as follows:
\begin{align}
	C_{xy}^{rec}=(\sum \lambda_{ij}U_{ij}-C'_{xy})^2
	\label{rewrite_rec}
\end{align}

\noindent According to the equation~\eqref{gradx}, consider the partial derivative of $C^{rec}_{xy}$ with respect to x in \eqref{rewrite_rec},
\begin{align}
	\frac{\partial C^{rec}_{xy}} {\partial x}=\sum 2(C_{xy}-C'_{xy})U_{ij} \frac{\partial \lambda_{ij}} {\partial x}
	\label{re_gx}
\end{align}
Similarly, the partial derivative with respect to y in \eqref{rewrite_rec}
\begin{align}
	\frac{\partial C^{rec}_{xy}} {\partial y}=\sum 2(C_{xy}-C'_{xy})U_{ij} \frac{\partial \lambda_{ij}} {\partial y}
	\label{re_gy}
\end{align}
If $\{U_{ij}\}$ are the same and the current coordinate $x,y$ fall into the center of an grid of 
sampling image, noticing that $ \{ \lambda_{ij} \} $ are weights of bilinear interpolation,  we have
\begin{align}
	\frac{\partial C^{rec}_{xy}} {\partial x}=0 \quad
	\frac{\partial C^{rec}_{xy}} {\partial y}=0
\end{align}
according to \eqref{re_gx}  and \eqref{re_gy}. This will cause a collapse in the coordinate learning for sampling. More specifically, the predicted coordinates can not be updated in the training due to its degenerate gradient.
To alleviate the issue, we adopt a decoder $D_{var}$ to apply variance to the sampling image. According to \eqref{re_gx}, 
\eqref{re_gy} and \eqref{grad_new_fn},  we can adjust $U_{ij}$ to $U'_{ij}$ with $I_{var}$ to optimize the gradient $\frac{\partial C^{rec}_{xy}} {\partial x}$ and $\frac{\partial C^{rec}_{xy}} {\partial y}$ for coordinates as follows, 
\begin{align}
	\frac{\partial C^{rec}_{xy}} {\partial x}&=\sum 2(C_{xy}-C'_{xy})U_{ij}I^{ij}_{var} \frac{\partial \lambda_{ij}} {\partial x}\\
	&=\sum 2(C_{xy}-C'_{xy})U'_{ij} \frac{\partial \lambda_{ij}} {\partial x}\\
	\frac{\partial C^{rec}_{xy}} {\partial y}&=\sum 2(C_{xy}-C'_{xy})U_{ij}I^{ij}_{var} \frac{\partial \lambda_{ij}} {\partial y}\\
	&=\sum 2(C_{xy}-C'_{xy})U'_{ij} \frac{\partial \lambda_{ij}} {\partial y}
	\label{last_gxy}
\end{align}
Different from previous work, the gradient for coordinates will be free from the collapse due to the regulator, $I_{var}$.

For $I_{flow}$ and $I_{var}$ learning, we adopt a pseudo-label, a general position for predicted coordinates, from SoftRas to optimize
$D_{flow}$ and $D_{var}$, preliminarily. The pseudo-label, however, only generates rough textures.
To further refine the texture flow, we additionally leverage image reconstruction to search reasonable region for coordinates. 
According to \eqref{grad_new_fn}, $I_{flow}$ and $I_{var}$ are adjusted to satisfy:

\begin{align}
	\min_{I_{var}, I_{flow}}  \| \theta_c^0+ \sum_{t} \alpha^{t} I_{grad}^{t}-Y_{pse}\|
\end{align}
and
\begin{align}
	\centering
	\min_{I_{var}, I_{flow}}   \|F_r(V_{pred},\phi(I_{uv}[\theta_c^0+ \sum_{t} \alpha^{t} I_{grad}^{t}]),cam)-I_r\|
\end{align}
$\theta^{0}_c$ are the initialization parameters and $Y_{pse}$ is the label providing the rough position for coordinates learning. $I^{t}_{grad}$ is the optimized gradient
$I^{new}_{grad}$ by $I_{var}$ at step t in the training.

\subsection{FID for Texture Learning}
For each texture, it can divided into two parts, including a visible part and an unseen part of an object.
 For convenience, we denote the visible part and unseen part as $T_{seen}$ and $T_{unseen}$, respectively. For each texture, we have,
\begin{align}
Tex=T_{seen} \cup T_{unseen}\\
\varnothing=T_{seen} \cap T_{unseen}
\end{align}
For $T_{seen}$, it can be predicted directly by image reconstruction as follows.
\begin{align}
	L_{rec}=\|F_{r}(V_{pred},T_{seen},cam)-I_{r}\|
\end{align}
Due to a lack of other views of an object, one can not predict the $T_{unseen}$ relying on image reconstruction. Fortunately, GAN, an efficient technique for unsupervised learning, provides weak supervised information for $T_{unseen}$ learning. More specifically, following the Differentiable Renderer, one can render the 3D Mesh with the texture from different views and obtain rendered images $I_{rand}$.
Then, $T_{unseen}$ is optimized to satisfy consistency between ground-true images and rendered images in terms of distribution. With adversarial learning, the texture will be of high fidelity from every view for 3D model.  

To learn parameters associated with $T_{unseen}$, we sample a camera pose $cam'$ randomly and obtain a rendered image from $cam'$.
In formula,
\begin{align}
	I_{rand}=F_r(V_{pred},Tex,cam')
\end{align}  
For the training of discriminator network $D$, it increases the photo-realism of predictions by rejecting image samples generated from our model. In formula,  
\begin{align}
	\max_{D} E_{x \sim P_{real}}[logD(x)]+E_{x \sim P_{I_{rand}}}[log(1-D(x))]
\end{align}  
 For the generator training, the rendered images should be accordant to input images in terms of distribution. 
  
\begin{align}
	\min_{I_{rand}} E_{x \sim P_{I_{rand}}}[log(1-D(x))]
\end{align}
To maintain stability of convergence in adversarial learning, we adopt a gradient penalty introduced from WGAN-GP\cite{gulrajani2017improved}.
\begin{align}
	L_{gp}=E_{x \sim \tilde{p}}[(\| \nabla_{x} D(x) \|-1)^2 ]
\end{align}
The images sampled from $\tilde{p}$ are the interpolation of samples from $P_{real}$ and $P_{I_{rand}}$.

\indent  Like most previous work, an distance function applied to feature helps
alleviate the domain gap between $P_{real}$ and $P_{I_{rand}}$ at a semantic level. 
 In our model, FID, an approach to measure the distance between two domain in feature space, is introduced to bridge the gap between 2D and 3D modals as a loss function. 
The inception model V3\cite{szegedy2016rethinking} pretrained on ImageNet\cite{deng2009imagenet}  is adopted to extract feature for FID.
Formally,
	

\begin{align}
    &f^{icp}_{real}=F_{icp}(I_{r})\\
    &f^{icp}_{rand}=F_{icp}(I_{rand})    	
\end{align}  
$F_{icp}$ is the feature extractor for input images and rendered images.  FID aims to calculates
the distance between two modals in terms of distribution according to  mean and covariance matrices,
assuming the feature $f^{icp}_{real}$ and $f^{icp}_{rand}$, generated from inception model, satisfy Gaussian distribution.
\begin{align}
L_{fid}=&||u_{real}-u_{pred} \|^2 \\
&+Tr(\sigma_{real}+\sigma_{pred}-2\sqrt{\sigma_{real} \sigma_{pred}})
\end{align} 
$u_{real}, \sigma_{real}$ are mean and covariance matrices of $f^{icp}_{real}$. 
 $u_{pred}$, $\sigma_{pred} $ are the mean and covariance matrices of feature $f^{icp}_{rand}$. Formally, for the mean calculation,
\begin{align}
	&u_{real}=\int_{x \sim P_{real}} F_{icp}(x)p(x)dx \\
	&u_{pred}=\int_{x \sim P_{I_{rand}}} F_{icp}(x)p(x)dx
\end{align}
For the covariance matrices calculation,
\begin{align}
	&\sigma_{real}=E_{x \sim {p_{real}}}[(x-u_{real})(x-u_{real})^{T}]\\
	&\sigma_{pred}=E_{x \sim {p_{pred}}}[(x-u_{pred})(x-u_{pred})^{T}]
\end{align}
$L_{fid}$ encourages our model to alleviate the domain gap between input images and rendered images from various views at a semantic level and regularizes the shape and texture learning.

\begin{figure*}[!t]
	\centering
	\includegraphics[width=0.8\linewidth]{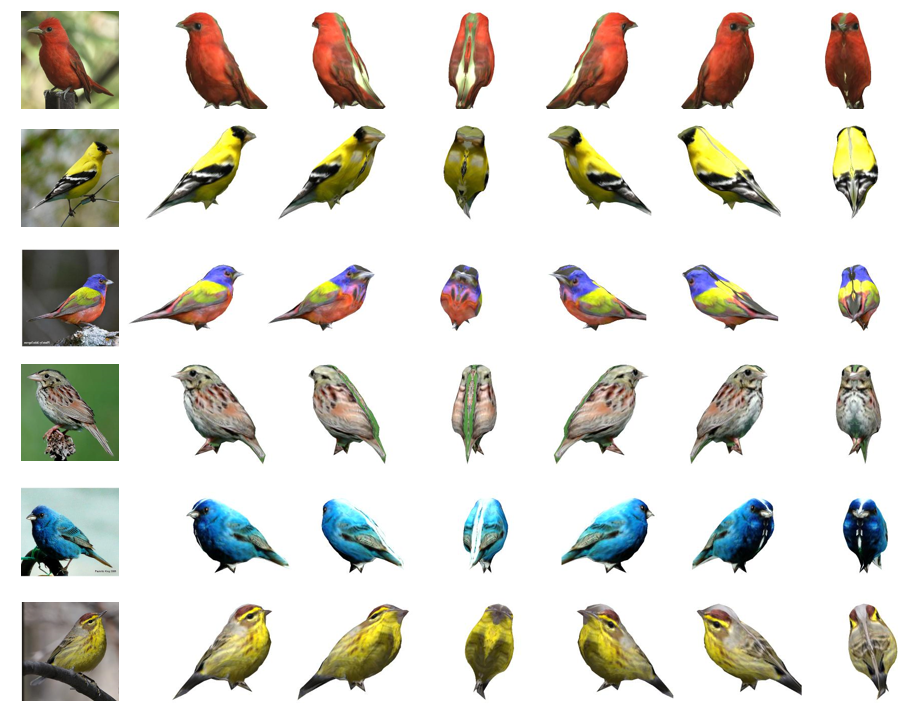}
	
	\caption{The first column is the input images. The second column
		is the rendered images from predicted camera pose. The camera pose learned by our model is
		almost close to the input image according to the first and second columns. 
        Rendered images from other views are illustrated from 
		the third to the last column. The results show that our model can generate the
		texture full of detail.
	} 
\label{ex0}
\end{figure*}

\begin{figure*}[!ht]
	\centering
	\includegraphics[width=0.8\linewidth]{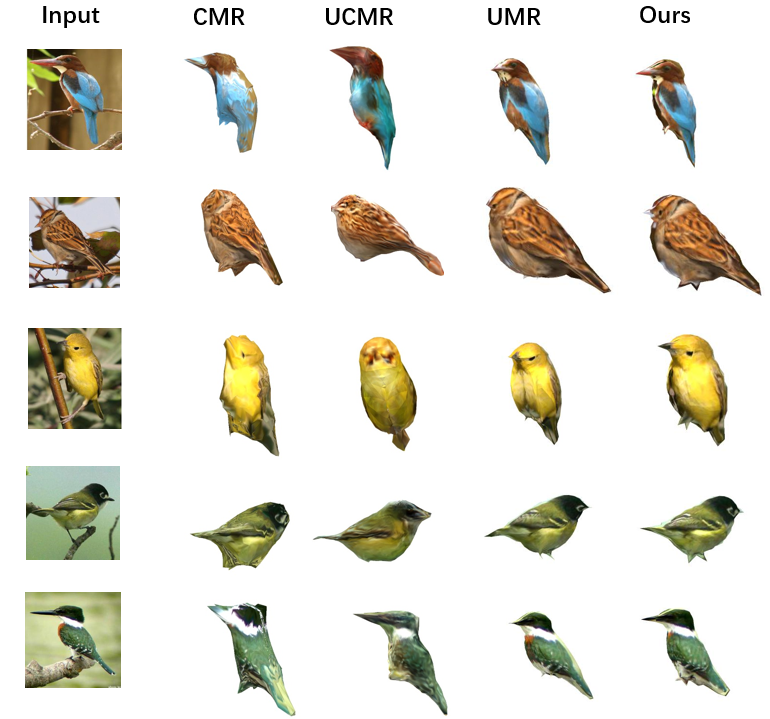}
	
	\caption{From the left to right, we present the input image, the results of CMR, UCMR, UMR, and our model. It can be observed that the shape from our model is more natural. Note, for example, the bill and legs. The texture generated by our algorithm also possesses more sharpness and detail, e.g., eyes and bill of the bird. To exhibit more detail and difference between the UMR and our model, we render some 3D models from various views in Figure \ref{ex_detail}. 		
	} 
\label{ex_db}
\end{figure*}

\begin{figure*}[!ht]
	\centering
	\includegraphics[width=0.8\linewidth]{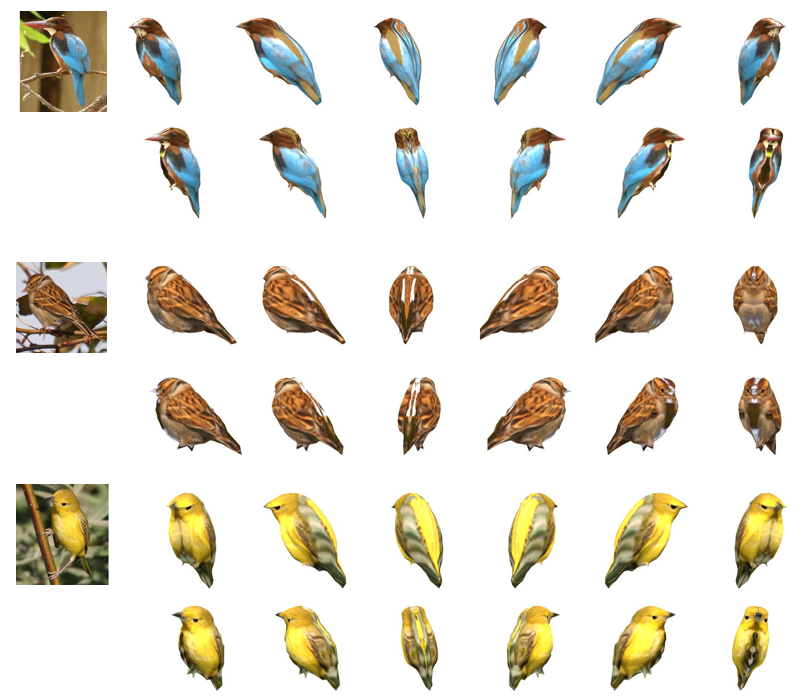}
	
	\caption{Comparing our model to UMR in three groups. In each group, the leftmost is the input image and the others are rendered images
		from various views(the top, UMR and the bottom, our model). As demonstrate in Figure \ref{ex_detail}, The 3D models by our method are full of detail, for example, the bill and the legs. For the region out of sight, the texture from our
		model is more reasonable, e.g., the wings and back.
	} 
	\label{ex_detail}
\end{figure*}

\section{Experiments}
\subsection{Experimental Settings}
\textbf{Network training:} In our experiments, the object generated from our algorithm is deformed from a sphere template, which has 642 vertices and 1280 faces. To maintain a stable convergence of the network, our training is divided into two stage. We first train a template for all instance  without $D_{var}$ and the loss function, FID\cite{li2020self}. 
 Based on the learned template, our model updates the deformation for the 3D shape and refine the texture flow with the decoder $D_{var}$ and and the loss function FID.

\noindent \textbf{Datasets: } We conduct experiments using only 2D images and their silhouettes and without 3D coordinates, 2D semantic keypoints or texture data. For each instance in our model, only a single-view image and its corresponding silhouette are treated as supervising information for geometry and texture generation. To refine each part of the object, semantic image segmentation is adopted in our model using SCOPS\cite{hung2019scops}.

\noindent \textbf{Pre-trained Model:} To speed up the convergence in shape and texture learning, the template for 
3D shape deformation is pre-trained according to UMR\cite{li2020self}. 
Image and vertex segmentation are included in the training to optimize the shape and texture in terms of details. Semantic image segmentation can be obtained by Hung\cite{hung2019scops} and vertices of 3D mesh can be labeled according to the relationship between the image with semantic label and vertices, established by the pretrained model.

\subsection{Quatitative Results}
\textbf{Single-view Reconstruction: } We compare our algorithm to other single-view 3D reconstruction methods\cite{li2020self,kanazawa2018learning,goel2020shape} presented recently. As showed in Figure \ref{ex_db}, We conduct our experiments on CUB datasets\cite{wah2011caltech}, the bird category. The resulting images are rendered from various views for visualization. Compared to previous work \cite{kanazawa2018learning,goel2020shape}, our model can achieve significant performance on each part of the object in shape reconstruction while maintaining the smoothness of the surface, and our textures are of higher fidelity and appear more photorealistic. In comparison with UMR \cite{li2020self}, the 3D model by our method retains more detail. Notice, for example, the bill and legs in Figure \ref{ex_detail}. These results demonstrate that the proposed variance decoder can optimize sampling coordinates to capture more  reasonable regions, the loss function, FID, can bridge the gap between a 2D image and a 3D model, thus improving geometry and texture learning.

\textbf{Texture Reconstruction} In our model, the variance decoder is adopted to optimize the texture by improving the gradient of predicted coordinates, and the FID loss function bridges the gap between rendered and ground-truth images for texture optimization at the semantic level. As shown in Figure \ref{ex0}, the texture flow learned by our model can search reasonable region for color sampling and accurately recover the fine details for 3D model, e.g. the eyes of the object.

\begin{table}
	\centering

	\begin{tabular}{ccc}
		\toprule
		Metric & Mask IOU& SSIM\\
		\midrule
		CMR&0.704& 0.782\\
		\midrule
		UCMR&0.6369&0.756\\
		\midrule
		UMR&0.734&0.812\\
		\midrule
		Ours & \textbf{0.7691} &  \textbf{ 0.8294}\\
	
		\bottomrule
	\end{tabular}
\caption{Evaluation for shape and texture. For shape, 2D IOU between predicted mask and ground true silhouette. Higher is better. SSIM evaluates the quality for texture generation. Higher is better.}
\label{iou}
\end{table}

\subsection{Quantitative Evaluations}

For shape evaluation, Due to the lack of ground-truth 3D shapes, we can not calculate the loss between predicted and ground-truth coordinates directly. Following Chen\cite{chen2019learning}, we compute the mask reprojection accuracy - the Intersection-Over-Union (IOU), between the rendered mask and the ground-truth silhouette. The comparisons on 2D IOU is shown in Table \ref{iou}. Our approach significantly outperforms other 3D unsupervised methods including CMR, UCMR and UMR, which demonstrates that our model achieves more consistent and photorealistic input images.

For texture, similar to shape evaluation, texture data is not included in our experiments. Although FID can
be adopted for texture evaluation, it is used as a loss function in our model and therefore excluded in
evaluation. Following the previous work by Oechsle\cite{oechsle2019texture}, we adopt the Structure Similarity Image Metric (SSIM)\cite{wang2004image} to estimate distance between the rendered image from a predicted camera pose and the ground-truth image. Since SSIM can capture local properties of an image, it can differentiate the details in the local region between the rendered image and the ground-truth counterpart. A quantitative comparison is shown
in Table \ref{iou}. One can see that our method achieves favorable performance than other methods, indicating that texture generation from our approach is more reasonable, corresponding to Figure \ref{ex_db}.

\section{Conclusion}
In this paper, we have proposed a novel framework for shape and texture learning to reconstruct 3D models only based on 2D supervision with single-view images. For texture learning, to prevent predicted
coordinates from being trapped into local regions and leading to collapse of sampling, we put forward a sampling algorithm to optimize gradient of learned coordinates by adopting a variance decoder to adjust the distribution of the sampling image. In addition, we introduce Fr$\acute{e}$chet Inception Distance (FID) as a loss function to bridge the gap between input images and rendered images. Synchronously, FID allows our model to enhance shape and texture learning for the unseen part of an object. The improvement of 3D models learned by our framework is confirmed by both objective metrics and visualization results.

\section{Reference}

\bibliographystyle{plain}
\bibliography{refer_own/refer}

\end{document}